%% file: sdgan.tex
\documentclass{article} 
\usepackage{iclr2018_conference,times}
\usepackage{hyperref}
\usepackage{url}

\usepackage[utf8]{inputenc} 
\usepackage[T1]{fontenc}    
\usepackage{hyperref}       
\usepackage{url}            
\usepackage{booktabs}       
\usepackage{amsfonts}       
\usepackage{amsmath}
\usepackage{nicefrac}       
\usepackage{microtype}      
\usepackage{graphicx}
\usepackage{subcaption}
\usepackage{amsmath}
\usepackage{bbm}
\usepackage{algorithm}
\usepackage[noend]{algpseudocode}
\usepackage[export]{adjustbox}
\usepackage{float}
\usepackage{bm}
\usepackage[multiple]{footmisc}

\newcommand{\vect}[1]{\mathbf{#1}}

\makeatletter
\newcommand{\shorteq}{%
  \settowidth{\@tempdima}{-}
  \resizebox{\@tempdima}{\height}{=}%
}
\makeatother

\title{Semantically Decomposing the Latent Spaces of Generative Adversarial Networks}

\author{Chris Donahue \\
Department of Music \\
University of California, San Diego \\
\texttt{cdonahue@ucsd.edu} \\
\And
Zachary C. Lipton \\
Carnegie Mellon University~~~~~~~~~~~~ \\
Amazon AI \\
\texttt{zlipton@cmu.edu} \\
\AND
Akshay Balsubramani \\
Department of Genetics \\
Stanford University \\
\texttt{abalsubr@stanford.edu} \\
\And
\hspace{37mm}Julian McAuley \\
\hspace{37mm}Department of Computer Science \\
\hspace{37mm}University of California, San Diego \\
\hspace{37mm}\texttt{jmcauley@eng.ucsd.edu} \\
}

\iclrfinalcopy

\begin{document}

\maketitle

\begin{abstract}
We propose a new algorithm for training 
generative adversarial networks 
that jointly learns latent codes
for both identities (e.g.~individual humans) 
and observations (e.g.~specific photographs).
By fixing the identity portion of
the latent codes,
we can generate diverse images of the same subject,
and by fixing the observation portion, 
we can traverse the manifold of subjects 
while maintaining contingent aspects
such as lighting and pose.
Our algorithm features a pairwise training scheme
in which each sample from the generator
consists of two images with a common identity code.
Corresponding samples from the real dataset consist
of two distinct photographs of the same subject.
In order to fool the discriminator, 
the generator must produce pairs
that are 
photorealistic, distinct,
and appear to depict the same individual. 
We augment both the DCGAN and BEGAN approaches 
with Siamese discriminators
to facilitate pairwise training.
Experiments with human judges 
and an off-the-shelf face verification system 
demonstrate our algorithm's ability 
to generate convincing, identity-matched photographs.
\end{abstract}

\section{Introduction}
\label{sec:introduction}
\input{sections/introduction.tex}

\section{Semantically Decomposed Generative Adversarial Networks }
\label{sec:methodology}
\input{sections/gans.tex}

\subsection{SD-GAN formulation}
\label{sec:sd-gans}
\input{sections/sd-gans.tex}

\section{Experiments}
\label{sec:experiments}
\input{sections/experiments.tex}

\subsection{Evaluation}
\label{sec:evaluation}
\input{sections/evaluation.tex}

\section{Related work}
\label{sec:related}
\input{sections/related.tex}

\section{Discussion}
\label{sec:discussion}
\input{sections/discussion.tex}

\input{sections/conclusions.tex}

\subsubsection*{Acknowledgements}
The authors would like to thank 
Anima Anandkumar, 
John Berkowitz 
and 
Miller Puckette 
for their helpful feedback on this work.
This work used the Extreme Science and Engineering Discovery Environment (XSEDE), which is supported by National Science Foundation grant number ACI-1053575~\citep{towns2014xsede}.
GPUs used in this research were donated by the NVIDIA Corporation.

\bibliography{sdgan}
\bibliographystyle{iclr2018_conference}

\clearpage
\appendix
\input{sections/appendix.tex}

\end{document}

%% file: sections/introduction.tex
In many domains, 
a suitable 
generative process might consist of several stages. 
To generate a photograph of a product, 
we might wish to first sample from the space of products, 
and then from the space of photographs \emph{of that product}. 
Given such disentangled representations 
in a multistage generative process, 
an online retailer might diversify its catalog, 
depicting products in a wider variety of settings. 
A retailer could also flip the process,
imagining new products in a fixed setting. 
Datasets for such domains often contain 
many labeled \emph{identities} with fewer \emph{observations} of each 
(e.g.~a collection of face portraits with thousands of people and ten photos of each). 
While we may know the identity of the subject in each photograph, 
we may not know the \emph{contingent} aspects of the observation 
(such as lighting, pose and background). 
This kind of data is ubiquitous; 
given a set of commonalities, 
we might want to incorporate this structure into our latent representations.

Generative adversarial networks (GANs)
learn mappings from latent codes $\vect{z}$ in some low-dimensional space $\mathcal{Z}$ 
to points in the space of natural data $\mathcal{X}$
\citep{goodfellow2014generative}. 
They achieve this power through an adversarial training scheme pitting a generative model $G : \mathcal{Z} \mapsto \mathcal{X}$ 
against a discriminative model $D : 
\mathcal{X} \mapsto [0,1]$ 
in a minimax game. 
While GANs 
are popular, owing to their 
ability to generate high-fidelity images, 
they do not, in their original form, 
explicitly disentangle the latent factors according to known commonalities. 

\textbf{In this paper}, we propose Semantically Decomposed GANs (SD-GANs), which encourage a specified portion of the latent space to correspond to a known source of variation.\footnote{Web demo: \url{https://chrisdonahue.github.io/sdgan}}\footnote{Source code: \url{https://github.com/chrisdonahue/sdgan}}
The technique decomposes
the latent code $\mathcal{Z}$
into one portion 
$\mathcal{Z}_I$ corresponding to identity,
and the remaining portion $\mathcal{Z}_O$ 
corresponding to the other contingent aspects of observations.
SD-GANs learn through a pairwise training scheme 
in which each sample from the real dataset 
consists of two distinct images with a common identity. 
Each sample from the generator 
consists of a pair of images
with common $\vect{z}_I \in \mathcal{Z}_I$ but differing $\vect{z}_O \in \mathcal{Z}_O$. 
In order to fool the discriminator,
the generator must not only produce 
diverse and photorealistic images,
but also images that depict the same identity 
when $\vect{z}_I$ is fixed. 
For SD-GANs,
we modify the discriminator 
so that it can determine 
whether a pair of samples constitutes a match.

\begin{figure}
  \centering
  \includegraphics[width=1.0\linewidth]{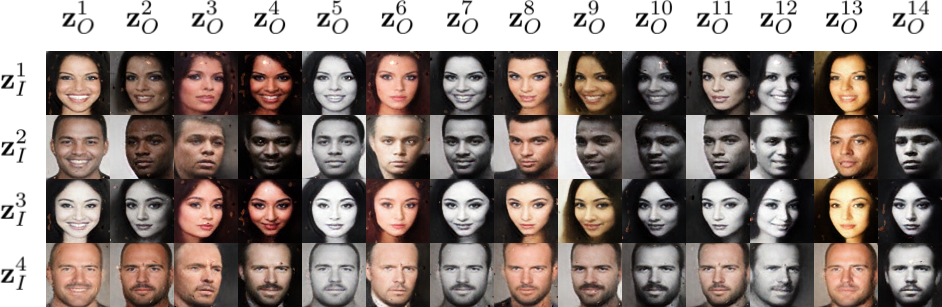}
  \caption{Generated samples from SD-BEGAN. Each of the four rows has the same identity code $\vect{z}_I$ 
and each of the fourteen columns has the same observation code $\vect{z}_O$.}
  \label{fig:began_celeb}
\end{figure}

As a case study, 
we experiment with a dataset of face photographs, 
demonstrating that SD-GANs can generate contrasting images 
of the same subject 
(Figure \ref{fig:began_celeb}; 
interactive web demo in footnote on previous page). 
The generator learns that certain properties 
are free to vary across observations but not identity. 
For example, SD-GANs learn 
that pose, facial expression, 
hirsuteness, 
grayscale vs.~color, 
and lighting 
can all vary across different photographs 
of the same individual.
On the other hand,
the aspects 
that are more salient for facial verification 
remain consistent as we vary the observation code $\vect{z}_O$. 
We also train SD-GANs 
on a dataset of product images,
containing multiple photographs of each product 
from various perspectives (Figure \ref{fig:sddcgan_shoes}).

We demonstrate that SD-GANs trained on faces
generate stylistically-contrasting, 
identity-matched image pairs 
that human annotators and a state-of-the-art face verification algorithm recognize 
as depicting the same subject. 
On measures of identity coherence and image diversity, SD-GANs perform comparably to a recent conditional GAN method~\citep{odena2016conditional}; 
SD-GANs can also imagine new identities, 
while conditional GANs are limited to generating existing identities from the training data. 

%% file: sections/gans.tex
Before introducing our algorithm, 
we briefly review the prerequisite concepts.

\subsection{GAN preliminaries}
GANs leverage the discriminative power of neural networks 
to learn generative models. 
The generative model $G$ ingests latent codes $\vect{z}$, 
sampled from some known prior $P_{\mathcal{Z}}$, 
and produces $G(\vect{z})$, 
a sample of an implicit distribution $P_G$. 
The learning process consists of a minimax game 
between $G$, 
parameterized by $\theta_G$,  
and a discriminative model $D$,
parameterized by $\theta_D$.
In the original formulation, 
the discriminative model tries to maximize log likelihood, 
yielding
\begin{equation}
\label{eq:gan}
\min_G \max_D V(G, D) = \mathbbm{E}_{\vect{x} \sim P_R}
[\log D(\vect{x})] + 
\mathbbm{E}_{\vect{z} \sim P_{\mathcal{Z}}}
[\log(1-D(G(\vect{z})))].
\end{equation}
Training proceeds as follows: 
For $k$ iterations, 
sample one minibatch 
from the real distribution $P_R$ 
and one from the distribution 
of generated images $P_G$, 
updating discriminator weights $\theta_D$ 
to increase $V(G,D)$ by stochastic gradient ascent.
Then sample a minibatch from $P_{\mathcal{Z}}$, 
updating $\theta_G$ to decrease $V(G,D)$ 
by stochastic gradient descent. 

\citet{zhao2016energy}
propose energy-based GANs (EBGANs),
in which the discriminator 
can be viewed as an energy function.
Specifically, they
devise a discriminator 
consisting of an autoencoder:
$D(\vect{x}) = D_{d}(D_{e}(\vect{x}))$.
In the minimax game, 
the discriminator's weights are updated 
to minimize the reconstruction error 
$\mathcal{L}(\vect{x}) = \vert \vert \vect{x} - D(\vect{x}) \vert \vert$ for real data, 
while maximizing the error $\mathcal{L}(G(\vect{z}))$ for the generator. 
More recently, \citet{berthelot2017began} extend this work, 
introducing Boundary Equilibrium GANs (BEGANs), 
which optimize the Wasserstein distance 
(reminiscent of Wasserstein GANs~\citep{arjovsky2017wasserstein}) 
between autoencoder loss distributions, yielding the formulation: 

\begin{equation}
\label{eq:began}
V_{BEGAN}(G, D) = \mathcal{L}(\vect{x}) - \mathcal{L}(G(\vect{z})).
\end{equation}

Additionally, they introduce a method for stabilizing training.
Positing that training becomes unstable
when the discriminator cannot distinguish 
between real and generated images,
they introduce a new hyperparameter $\gamma$,
updating the value function on each iteration to maintain a desired ratio between the two reconstruction errors:
$\mathbbm{E} [\mathcal{L}(G(\vect{z}))] = \gamma \mathbbm{E}[\mathcal{L}(\vect{x})]$.
The BEGAN model produces what appear to us, subjectively,
to be the sharpest images of faces 
yet generated by a GAN.
In this work, we adapt both the DCGAN~\citep{radford2015unsupervised} and BEGAN algorithms 
to the SD-GAN training scheme.

%% file: sections/sd-gans.tex
\begin{algorithm*}[t]
\caption{Semantically Decomposed GAN Training}\label{alg:identity-gan}
\begin{algorithmic}[1]
\For{n in 1:NumberOfIterations}
\For{m in 1:MinibatchSize}
	\State Sample one identity vector $\vect{z}_I \sim \text{Uniform}([-1,1]^{d_I})$.
    \State Sample two observation vectors 
    $\vect{z}^1_O, \vect{z}^2_O \sim \text{Uniform}([-1,1]^{d_O})$.
    \State $\vect{z}^1 \gets {[\vect{z}_I; \vect{z}^1_O]}$, $\vect{z}^2 \gets {[\vect{z}_I; \vect{z}^2_O]}$.
    \State Generate pair of images $G(\vect{z}^1), G(\vect{z}^2)$, adding them to the minibatch with label $0$ (fake).
\EndFor
\For{m in 1:MinibatchSize}
	\State Sample one identity $i \in \mathcal{I}$ uniformly at random from the real data set.
    \State Sample two images of $i$ without replacement $\vect{x}_1, \vect{x}_2 \sim P_R(\vect{x}|I = i)$.
    \State Add the pair to the minibatch, assigning label $1$ (real).
\EndFor
\State Update discriminator weights by 
$\theta_D \gets \theta_D + \nabla_{\theta_D} V(G,D)$
using its stochastic gradient.
\State Sample another minibatch of identity-matched latent vectors $\vect{z}^1, \vect{z}^2$.
\State Update generator weights by stochastic gradient descent
$\theta_G \gets \theta_G - \nabla_{\theta_G} V(G,D)$. 
\EndFor
\end{algorithmic}
\end{algorithm*}

Consider the data's identity as a random variable $I$ in a discrete index set $\mathcal{I}$. 
We seek to learn a latent representation
that conveniently decomposes the variation in the real data into two parts:
1) due to $I$, and
2) due to the other factors of variation in the data, packaged as a random variable $O$. 
Ideally, the decomposition of the variation in the data into $I$ and $O$ should correspond exactly to a decomposition of the latent space $\mathcal{Z} = \mathcal{Z}_I \times \mathcal{Z}_O$. 
This would permit convenient interpolation and other operations on the inferred subspaces $\mathcal{Z}_I$ and $\mathcal{Z}_O$.

A conventional GAN samples $I, O$ 
from their joint distribution. 
Such a GAN's generative model samples directly from an unstructured prior over the latent space.
It does not disentangle the variation in $O$ and $I$, for instance by modeling conditional distributions $P_G(O \mid I = i )$, 
but only models their average with respect to the prior on $I$. 

Our SD-GAN method learns such a latent space decomposition, 
partitioning the coordinates of $\mathcal{Z}$ 
into two parts representing the subspaces,  
so that any $\vect{z} \in \mathcal{Z}$ can be written as the concatenation $[\vect{z}_I;\vect{z}_O]$ of its identity representation $\vect{z}_I \in \mathbb{R}^{d_I} = \mathcal{Z}_I$ and its contingent aspect representation $\vect{z}_O \in \mathbb{R}^{d_O} = \mathcal{Z}_O$. 
SD-GANs achieve this through
a pairwise training scheme 
in which each sample from the real data
consists of $\vect{x}_1, \vect{x}_2 \sim P_R(\vect{x} \mid I=i)$,
a pair of images with a common identity $i \in \mathcal{I}$.
Each sample from the generator
consists of $G(\vect{z}_1), G(\vect{z}_2) \sim P_G(\vect{z} \mid  \mathcal{Z}_I=\vect{z}_I)$,
a pair of images generated from
a common identity vector $\vect{z}_I \in \mathcal{Z}_I$ 
but i.i.d. observation vectors $\vect{z}_O^1, \vect{z}_O^2 \in \mathcal{Z}_O$. 
We assign identity-matched pairs from $P_R$ the label $1$ 
and $\vect{z}_I$-matched pairs from $P_G$ the label $0$.
The discriminator can thus learn to reject pairs for either of two primary reasons: 
1) not photorealistic or 2) not plausibly depicting the same subject.
See Algorithm \ref{alg:identity-gan} 
for SD-GAN training pseudocode.

\begin{figure*}[t!]
	\centering
	\begin{subfigure}[b]{0.2\textwidth}
  		\centering
		\includegraphics[scale=.20]{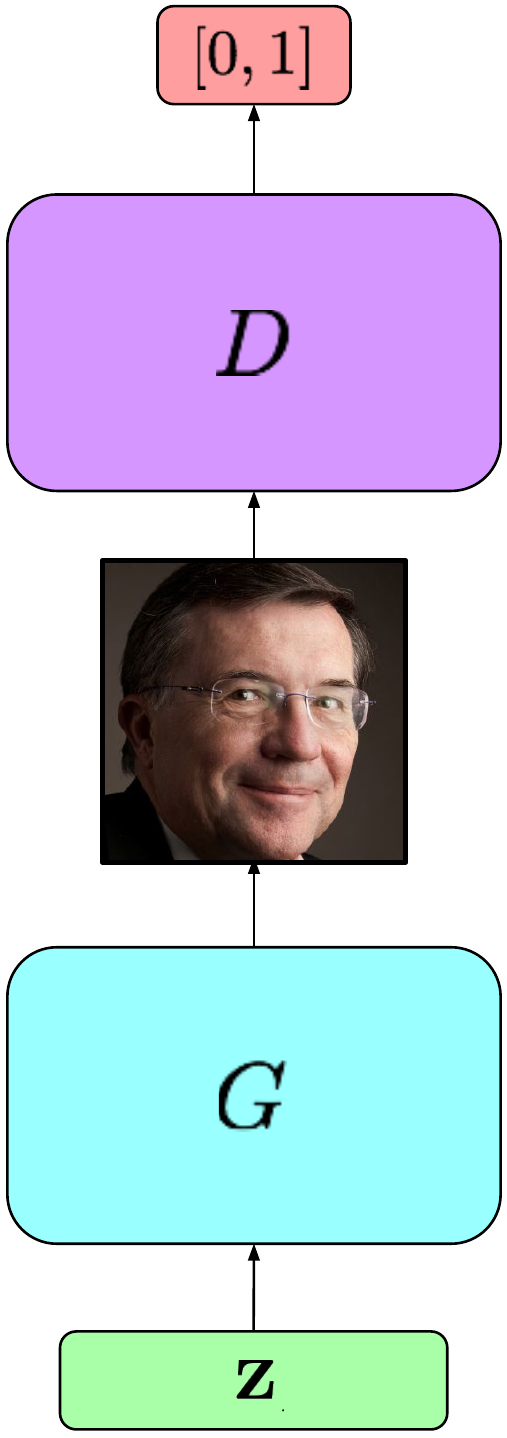}
        \vspace*{4mm}
        \caption{DCGAN}
        \label{fig:dcgan}
	\end{subfigure}~
    \begin{subfigure}[b]{0.3\textwidth}
        \centering
        \includegraphics[scale=.20]{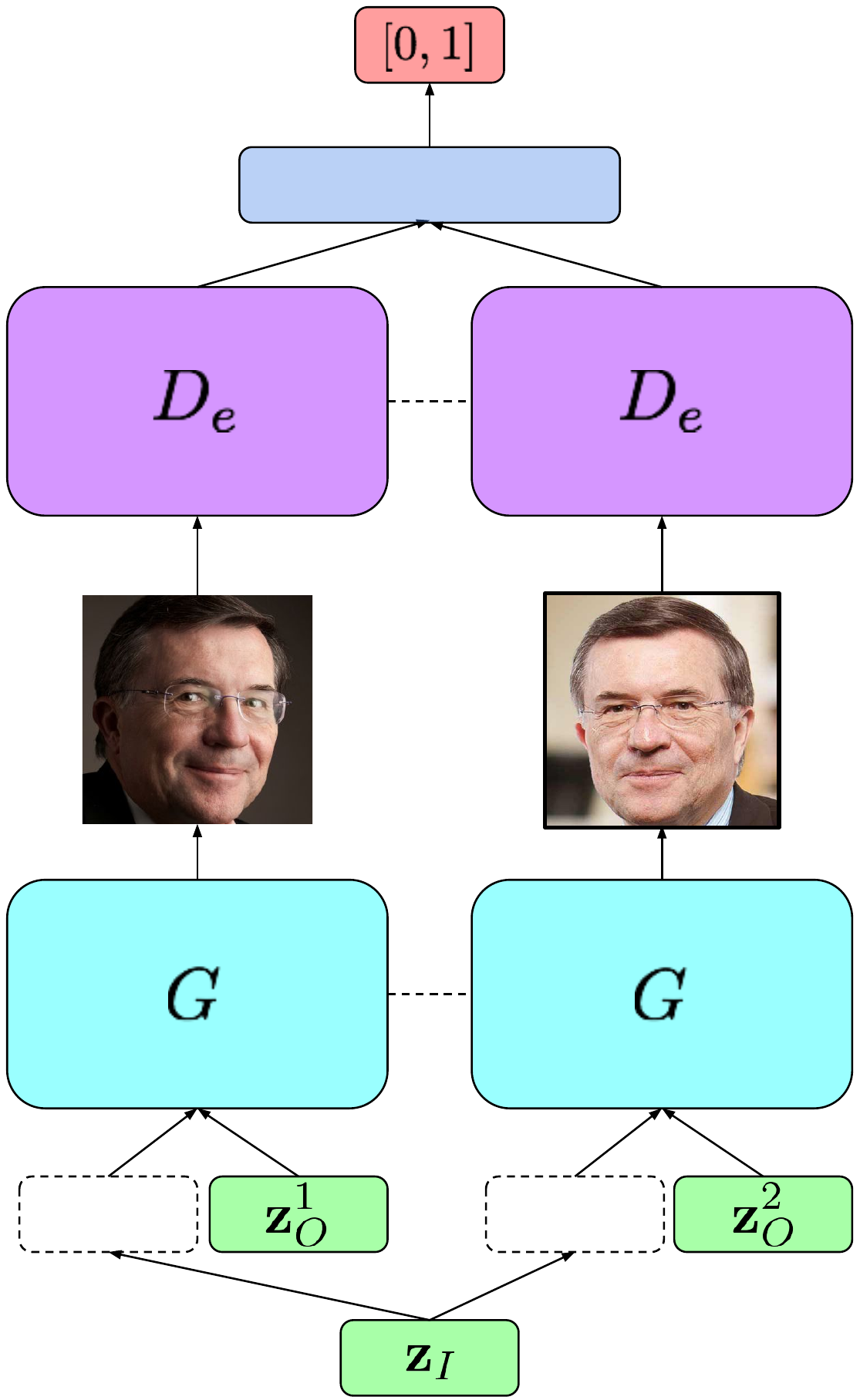}
        \caption{SD-DCGAN}
        \label{fig:sd-dcgan}
    \end{subfigure}~
	\begin{subfigure}[b]{0.2\textwidth} 
        \centering
        \includegraphics[scale=.20]{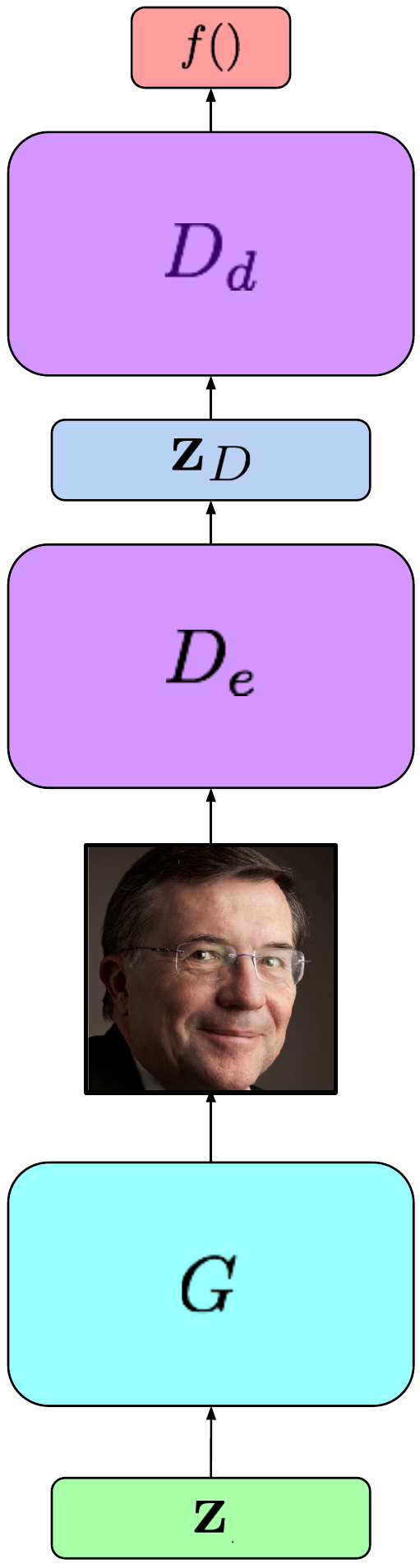}
        \vspace*{4mm}
        \caption{BEGAN}
        \label{fig:began}
	\end{subfigure}~
    \begin{subfigure}[b]{0.3\textwidth} 
  		\centering
		\includegraphics[scale=.20]{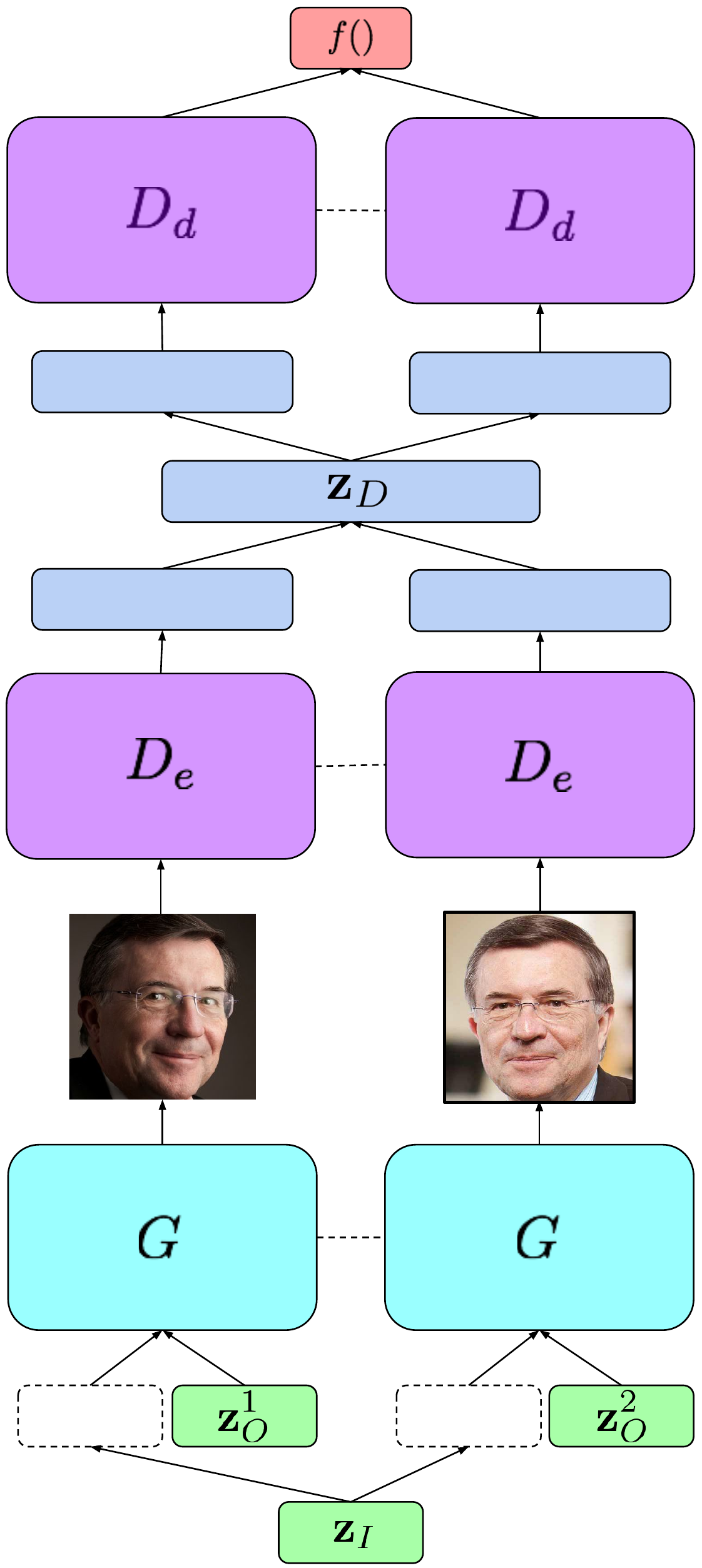}
		\caption{SD-BEGAN}
        \label{fig:sd-began}
	\end{subfigure}
	\caption{
SD-GAN architectures and vanilla counterparts. Our SD-GAN models incorporate a decomposed latent space and Siamese discriminators. Dashed lines indicate shared weights. Discriminators also observe real samples in addition to those from the generator (not pictured for simplicity).
}
\label{fig:architectures}
\end{figure*}

\subsection{SD-GAN discriminator architecture}
With SD-GANs, 
there is no need 
to alter the architecture of the generator. 
However, the discriminator must now act upon two images,
producing a single output.
Moreover, the effects of the two input images $\vect{x}_1, \vect{x}_2$
on the output score are not independent. 
Two images might be otherwise photorealistic but deserve rejection because they clearly depict different identities.  
To this end, we devise two novel discriminator architectures to adapt DCGAN and BEGAN respectively.
In both cases, we first separately encode each image using the same convolutional neural network $D_e$ (Figure~\ref{fig:architectures}).
We choose this Siamese setup \citep{bromley1993signature, chopra2005learning} as
our problem is symmetrical in the images,
and thus it's sensible to share weights between the encoders.

To adapt DCGAN,
we stack the feature maps $D_e(\vect{x}_1)$ and $D_e(\vect{x}_2)$ along the channel axis,
applying one additional strided convolution.
This allows the network to further aggregate information from the two images before flattening and fully connecting to a 
sigmoid output. 
For BEGAN, because the discriminator is an autoencoder, our architecture
is more complicated.
After encoding each image,
we concatenate the representations 
$[D_e(\vect{x}_1);D_e(\vect{x}_2)] \in \mathbb{R}^{2(d_I + d_O)}$ 
and apply one fully connected bottleneck layer $\mathbb{R}^{2(d_I + d_O)} \Rightarrow \mathbb{R}^{d_I + 2d_O}$
with linear activation.
In alignment with BEGAN,
the SD-BEGAN bottleneck has the same dimensionality as the tuple of latent codes 
($\vect{z}_I$, $\vect{z}^1_O$, $\vect{z}^2_O$)
that generated the pair of images.
Following the bottleneck, we apply a second FC layer
$\mathbb{R}^{d_I + 2d_O}\Rightarrow \mathbb{R}^{2(d_I + d_O)}$,
taking the first $d_I + d_O$ components of its output 
to be the input to the first decoder
and the second $d_I + d_O$ components 
to be the input to the second decoder.
The shared intermediate layer gives SD-BEGAN 
a mechanism to push apart matched and unmatched pairs.
We specify our exact architectures in full detail 
in Appendix~\ref{sec:appendix_arch}.

%% file: sections/experiments.tex
We experimentally validate SD-GANs using two datasets:
1) the \emph{MS-Celeb-1M} dataset of celebrity face images \citep{guo2016msceleb}
and 2) a dataset of shoe images collected from Amazon \citep{mcauley2015image}.
Both datasets contain a 
large number of identities
(people and shoes, respectively) 
with multiple observations of each. 
The ``in-the-wild'' nature of the celebrity face images offers a richer test bed for our method 
as both identities and contingent factors are significant sources of variation. 
In contrast, Amazon's shoe images tend to vary only with camera perspective for a given product, making this data useful for sanity-checking our approach. 

\begin{figure}
\centering
\begin{minipage}{.45\textwidth}
  \centering
  \includegraphics[width=1.0\linewidth]{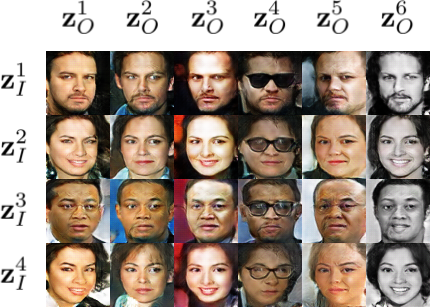}
  \captionof{figure}{Generated samples from \mbox{SD-DCGAN} model trained on faces.}
  \label{fig:sddcgan_celeb}
\end{minipage}\hfill
\begin{minipage}{.45\textwidth}
  \centering
  \includegraphics[width=1.0\linewidth]{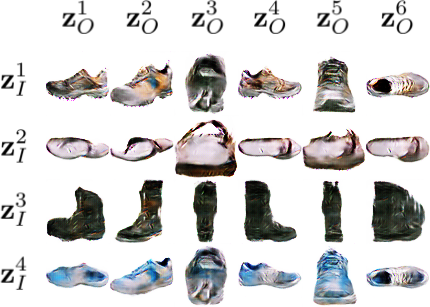}
  \captionof{figure}{Generated samples from \mbox{SD-DCGAN} model trained on shoes.}
  \label{fig:sddcgan_shoes}
\end{minipage}
\end{figure}

\paragraph{Faces}

From the aligned face images in the MS-Celeb-1M dataset, we select 12,500 celebrities at random
and $8$ associated images of each,
resizing them to $64$x$64$ pixels.
We split the celebrities into subsets of 10,000 (training), 1,250 (validation) and 1,250 (test).
The dataset has a small number of duplicate images
and some label noise 
(images matched to the wrong celebrity). 
We detect and remove duplicates by hashing the images, 
but we do not rid the data of label noise. 
We scale the pixel values to $[-1, 1]$, 
performing no additional preprocessing 
or data augmentation.

\paragraph{Shoes}
Synthesizing novel product images is another promising domain 
for our method.
In our shoes dataset, product photographs are captured 
against white backgrounds
and primarily differ in orientation and distance.
Accordingly, we expect that SD-GAN training 
will 
allocate the observation latent space 
to capture these aspects.
We choose to study shoes as a prototypical example of a category of product images.
The Amazon dataset 
contains around 3,000 unique products 
with the category ``Shoe'' and multiple product images.
We use the same $80\%, 10\%, 10\%$ split 
and again hash the images to ensure 
that the splits are disjoint.
There are $6.2$ photos of each product on average.

\subsection{Training details}
\label{sec:exp_deets}

We train SD-DCGANs on both of our datasets 
for 500,000 iterations 
using batches of $16$ identity-matched pairs.
To optimize SD-DCGAN, 
we use the Adam optimizer ~\citep{kingma2014adam} 
with hyperparameters 
$\alpha = 2\mathrm{e}{-4}, \beta_1 = 0.5, \beta_2 = 0.999$ 
as recommended by \citet{radford2015unsupervised}.
We also consider a non-Siamese discriminator 
that simply stacks the channels of the pair of real or fake images before encoding 
(SD-DCGAN-SC).

As in \citep{radford2015unsupervised}, 
we sample latent vectors $\vect{z} \sim \mbox{Uniform}([-1,1]^{100})$.
For SD-GANs, we 
partition the latent codes 
according to $\vect{z}_I \in \mathbb{R}^{d_I}, \vect{z}_O \in \mathbb{R}^{100 - d_I}$ using values of $d_I=[25, 50, 75]$.
Our algorithm can be trivially applied with $k$-wise training (vs.~pairwise). 
To explore the effects of using $k>2$,
we also experiment with an SD-DCGAN 
where we sample $k=4$ instances each from $P_G(\vect{z} \mid  \mathcal{Z}_I=\vect{z}_I)$
for some $\vect{z}_I \in \mathcal{Z}_I$
and from $P_R(\vect{x} \mid I=i)$ for some $i \in \mathcal{I}$.
For all experiments, unless otherwise stated, we use $d_I=50$ and $k=2$.

We also train an SD-BEGAN on both of our datasets. 
The increased complexity of the SD-BEGAN model significantly increases training time, 
limiting our ability to perform more-exhaustive hyperparameter validation (as we do for SD-DCGAN). 
We use the Adam optimizer with the default hyperparameters from \citep{kingma2014adam} for our SD-BEGAN experiments.
While results from our SD-DCGAN $k=4$ model are compelling, 
an experiment with a $k=4$ variant of SD-BEGAN resulted in early mode collapse (Appendix \ref{sec:honest_samples}); 
hence, we excluded SD-BEGAN $k=4$ from our evaluation.

We also compare to a DCGAN architecture 
trained using the auxiliary classifier GAN (AC-GAN) method~\citep{odena2016conditional}. 
AC-GAN differs from SD-GAN in two key ways: 
1) random identity codes $\vect{z}_I$ are replaced by a one-hot embedding over all the identities 
in the training set (matrix of size $10000$x$50$); 
2) 
the AC-GAN method encourages that generated photos depict the proper identity by
tasking its discriminator with predicting the identity of the generated or real image. 
Unlike SD-GANs, 
the AC-DCGAN model cannot imagine new identities;
when generating from AC-DCGAN 
(for our quantitative comparisons to SD-GANs), 
we must sample a random identity 
from those existing in the training data. 

%% file: sections/evaluation.tex
The evaluation of generative models 
is a fraught topic. 
Quantitative measures of sample quality 
can be poorly correlated with each other~\citep{theis2015note}.
Accordingly, we design an evaluation to match conceivable uses of our algorithm. 
Because we hope to produce diverse samples 
that humans deem to depict the same person,
we evaluate the identity coherence of SD-GANs 
and baselines 
using both a pretrained face verification model 
and crowd-sourced human judgments 
obtained through Amazon's Mechanical Turk platform. 

\subsubsection{Quantitative}

Recent advancements in face verification 
using deep convolutional neural networks \citep{schroff2015facenet,parkhi2015deep,wen2016discriminative} 
have yielded accuracy rivaling humans.
For our evaluation, 
we procure \emph{FaceNet},
a publicly-available face verifier 
based on the Inception-ResNet architecture \citep{szegedy2016inception}.
The \emph{FaceNet} model was pretrained 
on the CASIA-WebFace dataset \citep{yi2014learning}
and achieves $98.6\%$ accuracy 
on the LFW benchmark \citep{huang2012learning}.\footnote{``20170214-092102'' pretrained model from \url{https://github.com/davidsandberg/facenet}}

FaceNet ingests normalized, $160$x$160$ color images and produces an embedding $f(\vect{x}) \in \mathbb{R}^{128}$.
The training objective for FaceNet 
is to learn embeddings 
that minimize the $L_2$ distance 
between matched pairs of faces 
and maximize the distance for mismatched pairs.
Accordingly, the embedding space 
yields a function for measuring
the similarity between two faces $\vect{x}_1$ and $\vect{x}_2$: 
$D(\vect{x}_1, \vect{x}_2) = ||f(\vect{x}_1) - f(\vect{x}_2)||_{2}^{2}$.
Given two images, $\vect{x}_1$ and $\vect{x}_2$,
we label them as a match if 
$D(\vect{x}_1, \vect{x}_2) \leq \tau_{v}$ 
where $\tau_{v}$ is the 
accuracy-maximizing threshold 
on a class-balanced set of pairs from
MS-Celeb-1M validation data.
We use the same threshold for
evaluating both real and synthetic data
with FaceNet.

\begin{table}[!t]
  \centering
  \caption{
    Evaluation of $10$k pairs from \emph{MS-Celeb-1M} (real data) and generative models; 
    half have matched identities, half do not. 
    The \emph{identity verification} metrics demonstrate that FaceNet (FN) and human annotators on Mechanical Turk (MT) verify generated data similarly to real data. 
    The \emph{sample diversity} metrics ensure that generated samples are statistically distinct in pixel space. 
    Data generated by our best model (SD-BEGAN) performs comparably to real data.
    * $1$k pairs, \textdagger{} $200$ pairs.
  }

  \footnotesize
  \begin{tabular}{lr|crrr|rr}
    \multicolumn{2}{c}{} & \multicolumn{4}{c}{\emph{Identity Verification}} & \multicolumn{2}{c}{\emph{Sample Diversity}} \\
    \toprule
Dataset&Mem&Judge&AUC&Acc.&FAR&ID-Div&All-Div\\
\midrule
\emph{MS-Celeb-1M}&-&FN&$.913$&$.867$&$.045$&$.621$&$.699$\\
AC-DCGAN&$131$~MB&FN&$.927$&$.851$&$\mathbf{.083}$&$.497$&$.666$\\
SD-DCGAN&$57$~MB&FN&$.823$&$.749$&$.201$&$.521$&$.609$\\
SD-DCGAN-SC&$\mathbf{47}$~MB&FN&$.831$&$.757$&$.180$&$.560$&$.637$\\
SD-DCGAN $k\shorteq{}4$&$75$~MB&FN&$.852$&$.776$&$.227$&$.523$&$.614$\\
SD-DCGAN $d_I\shorteq{}25$&$57$~MB&FN&$.835$&$.764$&$.222$&$.526$&$.615$\\
SD-DCGAN $d_I\shorteq{}75$&$57$~MB&FN&$.816$&$.743$&$.268$&$.517$&$.601$\\
SD-BEGAN&$68$~MB&FN&$\mathbf{.928}$&$\mathbf{.857}$&$.110$&$\mathbf{.588}$&$\mathbf{.673}$\\
\midrule
\textdagger{}\emph{MS-Celeb-1M}&-&Us&-&$.850$&$.110$&$.621$&$.699$\\
*\emph{MS-Celeb-1M}&-&MT&-&$.759$&$.035$&$.621$&$.699$\\
*AC-DCGAN&$131$~MB&MT&-&$\mathbf{.765}$&$\mathbf{.090}$&$.497$&$.666$\\
*SD-DCGAN $k\shorteq{}4$&$75$~MB&MT&-&$.688$&$.147$&$.523$&$.614$\\
*SD-BEGAN&$\mathbf{68}$~MB&MT&-&$.723$&.096&$\mathbf{.588}$&$\mathbf{.673}$\\
    \bottomrule
  \end{tabular}

  \label{facepairs}
\end{table}

We compare the performance of FaceNet
on pairs of images from the MS-Celeb-1M test set against generated samples from our trained SD-GAN models and AC-DCGAN baseline. 
To match FaceNet's training data, 
we preprocess all images 
by resizing from $64$x$64$ to $160$x$160$, 
normalizing each image individually.
We prepare 10,000 pairs from MS-Celeb-1M, 
half identity-matched and half unmatched.
From each generative model, we generate 5,000 pairs each with $\vect{z}_I^1 = \vect{z}_I^2$ and 5,000 pairs with $\vect{z}_I^1 \neq \vect{z}_I^2$.
For each sample, we draw observation vectors $\vect{z}_O$ randomly.

We also want to ensure that identity-matched images produced by the generative models are diverse.
To this end, we propose an intra-identity sample diversity (ID-Div) metric. 
The multi-scale structural similarity (MS-SSIM)~\citep{wang2003multiscale} metric 
reports the similarity of two images on a scale from $0$ (no resemblance) to $1$ (identical images). 
We report $1$ minus the mean MS-SSIM for all pairs of identity-matched images as ID-Div. 
To measure the \emph{overall} sample diversity (All-Div), 
we also compute $1$ minus the mean similarity 
of $10$k pairs with random identities. 

In Table~\ref{facepairs}, 
we report the area under the receiver operating characteristic curve (AUC), 
accuracy, and false accept rate (FAR) of FaceNet (at threshold $\tau_{v}$) on the real and generated data. 
We also report our proposed diversity statistics. 
FaceNet verifies pairs from the real data with $87\%$ accuracy
compared to $86\%$ on pairs from our SD-BEGAN model. 
Though this is comparable to the accuracy achieved on pairs from the AC-DCGAN baseline, 
our model produces samples that are more diverse in pixel space (as measured by ID-Div and All-Div). 
FaceNet has a higher but comparable FAR for pairs from SD-GANs than those from AC-DCGAN; 
this indicates that SD-GANs may produce images that are less \emph{semantically} diverse on average than AC-DCGAN.

We also report the combined memory footprint of $G$ and $D$ for all methods in Table~\ref{facepairs}. 
For conditional GAN approaches, 
the number of parameters grows linearly
with the number of identities in the training data. 
Especially in the case of the AC-GAN, 
where the discriminator computes a softmax over all identities, 
linear scaling may be prohibitive. 
While our $10$k-identity subset of MS-Celeb-1M requires a $131$MB AC-DCGAN model, 
an AC-DCGAN for all $1$M identities would be over $8$GB, 
with more than $97$\% of the parameters devoted to the weights in the discriminator's softmax layer. 
In contrast, 
the complexity of SD-GAN is constant in the number of identities. 

\subsubsection{Qualitative}

In addition to validating 
that identity-matched SD-GAN samples are verified by FaceNet,
we also demonstrate that humans 
are similarly convinced
through experiments using Mechanical Turk.
For these experiments, 
we use balanced subsets of 1,000 pairs 
from MS-Celeb-1M and the most promising generative methods from our FaceNet evaluation. 
We ask human annotators to determine 
if each pair depicts 
the ``same person'' or ``different people''.
Annotators are presented with batches of ten pairs  at a time.
Each pair is presented to three distinct annotators and predictions are determined by majority vote. 
Additionally, to provide a benchmark 
for assessing the quality
of the Mechanical Turk ensembles,
we (the authors) manually judged $200$ pairs from MS-Celeb-1M. 
Results are in Table~\ref{facepairs}.

For all datasets, human annotators on Mechanical Turk
answered ``same person'' less frequently than FaceNet when the latter uses the accuracy-maximizing threshold $\tau_v$. 
Even on real data, 
balanced so that $50\%$ of pairs are identity-matched, 
annotators report ``same person'' only $28\%$ of the time (compared to the $41\%$ of FaceNet). 
While annotators achieve higher accuracy on pairs from AC-DCGAN than pairs from SD-BEGAN, 
they also answer ``same person'' $16\%$ more often for AC-DCGAN pairs than real data. 
In contrast, annotators answer ``same person'' 
at the same rate for SD-BEGAN pairs as real data. 
This may be attributable to the lower sample diversity produced by AC-DCGAN. 
Samples from SD-DCGAN 
and SD-BEGAN 
are shown in Figures \ref{fig:sddcgan_celeb} 
and \ref{fig:began_celeb} respectively. 

%% file: sections/related.tex
Style transfer
and novel view synthesis
are active research areas.
Early attempts to disentangle style and content manifolds
used factored tensor representations~\citep{tenenbaum1997separating,vasilescu2002multilinear,elgammal2004separating,tang2013tensor},
applying their results to 
face image synthesis.
More recent work focuses on learning hierarchical feature representations using deep convolutional neural networks to separate identity and pose manifolds for
faces~\citep{zhu2013deep, reed2014learning,zhu2014multi,yang2015weakly,kulkarni2015deep,oord2016conditional,yan2016attribute2image} 
and products~\citep{dosovitskiy2015learning}.
\citet{gatys2016image} use features of a convolutional network, pretrained for image recognition, as a means for discovering content and style vectors.

Since their introduction~\citep{goodfellow2014generative}, 
GANs have been used to generate increasingly
high-quality images \citep{radford2015unsupervised,zhao2016energy,berthelot2017began}.
Conditional GANs (cGANs), introduced by \citet{mirza2014conditional}, extend GANs 
to generate class-conditional data. 
\citet{odena2016conditional} propose auxiliary classifier GANs, 
combining cGANs with a semi-supervised discriminator~\citep{springenberg2015unsupervised}. 
Recently,
cGANs have been used 
to ingest text~\citep{reed2016generative} and 
full-resolution images~\citep{isola2016image,liu2017unsupervised,zhu2017unpaired} 
as conditioning information, 
addressing a variety of image-to-image
translation and style transfer tasks.
\citet{chen2016info} devise an information-theoretic extension to GANs 
in which they maximize the mutual information between a subset of latent variables and the generated data. 
Their unsupervised method appears to disentangle some intuitive factors of variation, 
but these factors may not correspond to those explicitly disentangled by SD-GANs.

Several related papers use GANs for 
novel view synthesis of faces. 
\citet{tran2017disentangled,huang2017beyond,yin2017semi,yin2017towards,zhao2017multi} 
all address synthesis of different body/facial poses conditioned on an input image (representing identity) and a fixed number of pose labels. 
\citet{antipov2017face} propose conditional GANs
for synthesizing artificially-aged faces 
conditioned on both a face image and an age vector. 
These approaches all require \emph{explicit} conditioning on the relevant factor 
(such as rotation, lighting and age)
in addition to an identity image. 
In contrast, SD-GANs can model these contingent factors implicitly (without supervision). 

\citet{mathieu2016disentangling} combine GANs with a traditional reconstruction loss to disentangle identity. 
While their approach trains 
with an encoder-decoder generator, 
they enforce a variational bound 
on the encoder embedding, 
enabling them to sample from the decoder
without an input image. 
Experiments with their method only address 
small ($28$x$28$) grayscale face images, 
and their training procedure is complex to reproduce. 
In contrast, our work offers a simpler approach 
and can synthesize higher-resolution, 
color photographs.

One might think of our work as offering 
the generative view of the Siamese networks
often favored for learning similarity metrics
 \citep{bromley1993signature, chopra2005learning}. 
Such approaches are used for discriminative tasks 
like face or signature verification 
that share the \emph{many classes with few examples} structure 
that we study here.
In our work, we adopt a Siamese architecture in order to enable the discriminator to differentiate between matched and unmatched pairs.
Recent work by \citet{liu2016coupled} propose a GAN architecture with weight sharing across multiple generators and discriminators,
but with a different problem formulation 
and objective from ours.

%% file: sections/discussion.tex
\begin{figure*}[t!]
    \centering
    \begin{subfigure}[b]{0.27\textwidth}
        \centering
        \includegraphics[width=1.0\linewidth]{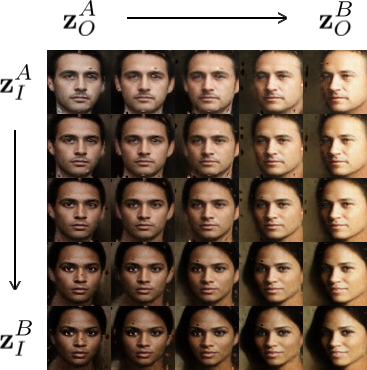}
    \end{subfigure}
    \hfill
    \begin{subfigure}[b]{0.27\textwidth}
        \centering
        \includegraphics[width=1.0\linewidth]{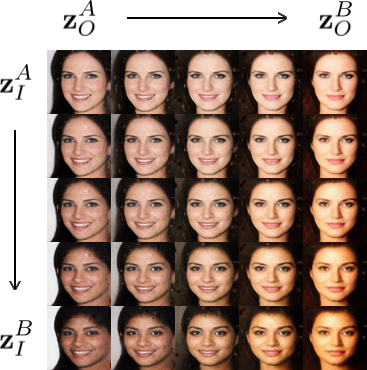}
    \end{subfigure}
    \hfill
    \begin{subfigure}[b]{0.27\textwidth}
        \centering
        \includegraphics[width=1.0\linewidth]{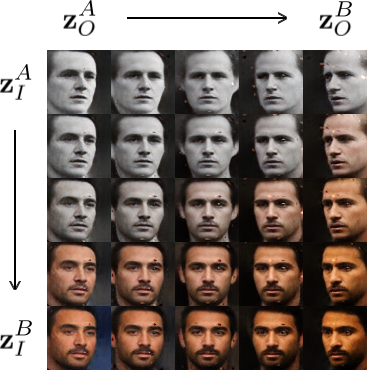}
    \end{subfigure}
    \caption{
    Linear interpolation of $\vect{z}_I$ (identity) and $\vect{z}_O$ (observation) for three pairs using SD-BEGAN generator. 
    In each matrix, rows share $\vect{z}_I$ and columns share $\vect{z}_O$.}
\label{fig:zplots}
\end{figure*}

Our evaluation demonstrates that SD-GANs
can disentangle 
those 
factors of variation corresponding to identity from the rest.
Moreover, with SD-GANs we can sample never-before-seen identities, a benefit not shared by 
conditional GANs. 
In Figure \ref{fig:sddcgan_celeb}, we demonstrate that by varying the observation vector $\vect{z}_O$,
SD-GANs can change the color of clothing, 
add or remove sunnies, or change facial pose. 
They can also perturb the lighting, color saturation, and contrast of an image,
all while keeping the apparent identity fixed. 
We note, subjectively, that samples from SD-DCGAN 
tend to appear less photorealistic than those from SD-BEGAN. 
Given a generator trained with SD-GAN, 
we can independently interpolate along 
the identity and observation manifolds (Figure \ref{fig:zplots}).

On the shoe dataset, we find that the SD-DCGAN model produces convincing results.
As desired, manipulating $\vect{z}_I$ while keeping $\vect{z}_O$ fixed 
yields distinct shoes in consistent poses (Figure \ref{fig:sddcgan_shoes}).
The identity code $\vect{z}_I$ appears to capture the broad categories of shoes 
(sneakers, flip-flops, boots, etc.).
Surprisingly, neither original BEGAN
nor SD-BEGAN can produce diverse shoe images (Appendix~\ref{sec:appendix_shoes}).

%% file: sections/conclusions.tex
In this paper, we presented SD-GANs, 
a new algorithm capable of disentangling factors of variation according to known commonalities.
We see several promising directions for future work.
One logical extension is to disentangle latent factors corresponding to more than one known commonality.
We also plan to apply our approach in other domains such as 
identity-conditioned speech synthesis.

%% file: sections/appendix.tex
\section{Estimating latent codes}

\begin{figure*}[ht!]
	\centering
	\begin{subfigure}[b]{0.45\textwidth} 
  		\centering
		\includegraphics[scale=0.40]{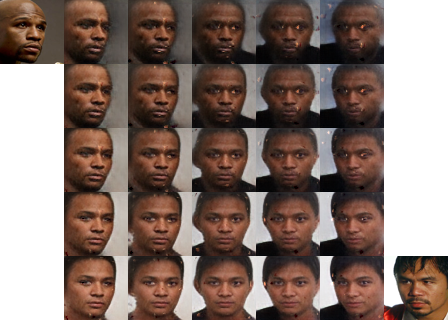}
	\end{subfigure}~
    \begin{subfigure}[b]{0.45\textwidth} 
  		\centering
		\includegraphics[scale=0.40]{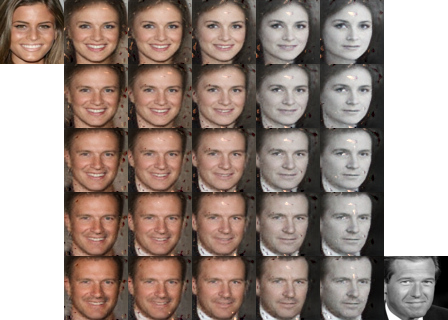}
	\end{subfigure}
	\caption{Linear interpolation of both identity (vertical) and observation (horizontal) on latent codes recovered for unseen images. All rows have the same identity vector ($\vect{z}_I$) and all columns have the same observation vector ($\vect{z}_O$).}
\label{fig:cyclone}
\end{figure*}

We estimate latent vectors for unseen images and demonstrate that the disentangled representations of SD-GANs can be used to depict the estimated identity with different contingent factors. In order to find a latent vector $\hat{\vect{z}}$ such that $G(\hat{\vect{z}})$ (pretrained $G$) is similar to an unseen image $\vect{x}$, 
we can minimize the distance between $\vect{x}$ and $G(\hat{\vect{z}})$: 
$\min_{\hat{\vect{z}}} ||G(\hat{\vect{z}}) - \vect{x}||_2^2$~\citep{lipton2017precise}. 

In Figure \ref{fig:cyclone}, we depict estimation and linear interpolation across both subspaces for two pairs of images using SD-BEGAN.
We also display the corresponding source images being estimated.
For both pairs, 
$\hat{\vect{z}}_I$ (identity) is consistent in each row and $\hat{\vect{z}}_O$ (observation) is consistent in each column.

\section{Pairwise discrimination of embeddings and encodings}
\label{sec:appendix_sdgan_embed}

\begin{figure}[H]
  \centering
  \includegraphics[width=0.96\linewidth]{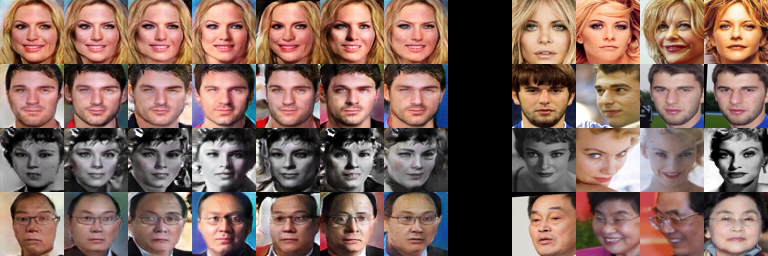}
  \caption{Generator with a one-hot identity embedding trained against a pairwise discriminator. Each row shares an identity vector and each column shares an observation vector. Random sample of $4$ real images of the corresponding identity on the right.}
  \label{fig:sdgan_embed}
\end{figure}

In Section~\ref{sec:exp_deets}, 
we describe an AC-GAN~\citep{odena2016conditional} baseline which uses an embedding matrix over real identities as latent identity codes ($G : {i, \vect{z}_O} \mapsto \vect{\hat{x}}$). 
In place of random identity vectors, we tried combining this identity representation with pairwise discrimination (in the style of SD-GAN). 
In this experiment, 
the discriminator receives either either two real images with the same identity $(\vect{x}_i^1, \vect{x}_i^2)$, or a real image with label $i$ and synthetic image with label $i$ $(\vect{x}_i^1, G(i, \vect{z}_O))$. 
All other hyperparameters are the same as in our SD-DCGAN experiment (Section~\ref{sec:exp_deets}). 
We show results in Figure~\ref{fig:sdgan_embed}. 

\begin{figure}[H]
  \centering
  \includegraphics[width=0.96\linewidth]{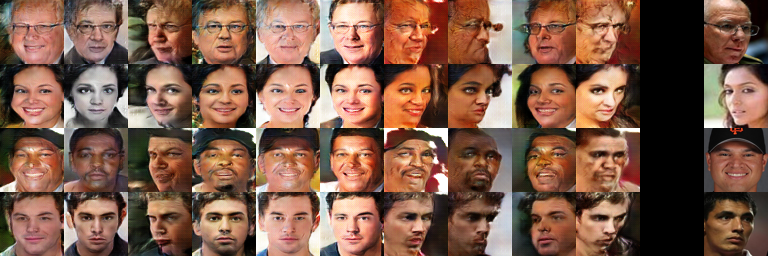}
  \caption{Generator with an encoder-decoder architecture trained against a pairwise discriminator. Each row shares an identity vector and each column shares an observation vector. Input image on the right.}
  \label{fig:sdgan_encode}
\end{figure}

In Appendix~\ref{sec:appendix_tran}, 
we detail a modification of the DR-GAN~\citep{tran2017disentangled} method which uses an encoding network $G_e$ to transform images to identity representations ($G_{d} : {G_{e}(\vect{x}), \vect{z_O}} \mapsto \vect{\hat{x}}$). 
We also tried combining this encoder-decoder approach with pairwise discrimination. 
The discriminator receives either two real images with the same identity $(\vect{x}_i^1, \vect{x}_i^2)$, 
or ($\vect{x}_i^1$, $G_{d}(G_{e}(\vect{x}_i^1), \vect{z}_O)$. 
We show results in Figure~\ref{fig:sdgan_encode}.

While these experiments are exploratory and not part of our principle investigation, 
we find the results to be qualitatively promising. 
We are not the first to propose pairwise discrimination with pairs of (real, real) or (real, fake) images in GANs~\citep{pathak2016context,isola2016image}.

\section{Exploratory experiment with DR-GANs}
\label{sec:appendix_tran}

\citet{tran2017disentangled} propose \emph{Disentangled Representation learning-GAN} (DR-GAN), an approach to face frontalization with similar setup to our SD-GAN algorithm. The (single-image) DR-GAN generator $G$ (composition of $G_{e}$ and $G_{d}$) accepts an input image $\vect{x}$, a pose code $\vect{c}$, and a noise vector $\vect{z}$. The DR-GAN discriminator receives either $\vect{x}$ or $\vect{\hat{x}} = G_{d}(G_{e}(\vect{x}), \vect{c}, \vect{z})$. In the style of~\citep{springenberg2015unsupervised}, the discriminator is tasked with determining not only if the image is real or fake, but also classifying the pose $\vect{c}$, suggesting a disentangled representation to the generator. Through their experiments, they demonstrate that DR-GAN can explicitly disentangle pose and illumination ($\vect{c}$) from the rest of the latent space ($G_{e}(\vect{x});\vect{z}$).

\begin{figure}[h]
  \centering
  \includegraphics[width=0.96\linewidth]{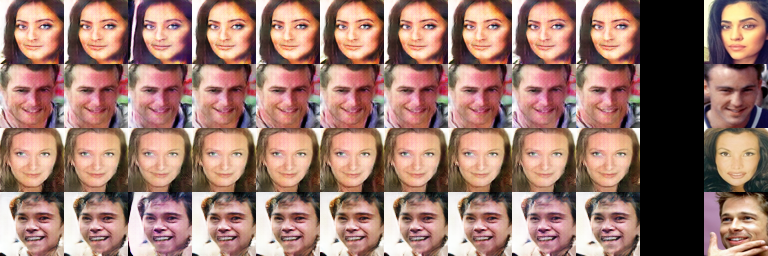}
  \caption{Generated samples from cGAN trained only to disentangle identity. Each row shares an identity vector and each column shares an observation vector; input image on the right.}
  \label{fig:tran_drgan}
\end{figure}

In addition to our AC-DCGAN baseline~\citep{odena2016conditional}, we tried modifying DR-GAN to only disentangle identity (rather than \emph{both} identity and pose in the original paper). We used the DCGAN~\citep{radford2015unsupervised} discriminator architecture (Table~\ref{tab:dcgan_disc}) as $G_{e}$, linearly projecting the final convolutional layer to $G_{e}(\vect{x}) \in \mathbb{R}^{50}$ (in alignment with our SD-GAN experiments). We altered the discriminator to predict the identity of $\vect{x}$ or $\vect{\hat{x}}$, rather than pose information (which is unknown in our experimental setup). With these modifications, $G_{e}(\vect{x})$ is analogous to $\vect{z_I}$ in the SD-GAN generator, and $\vect{z}$ is analogous to $\vect{z_O}$. Furthermore, this setup is identical to the AC-DCGAN baseline except that the embedding matrix is replaced by an encoding network $G_{e}$. Unfortunately, we found that the generator quickly learned to produce a single output image $\vect{\hat{x}}$ for each input $\vect{x}$ regardless of observation code $\vect{z}$ (Figure~\ref{fig:tran_drgan}). Accordingly, we excluded this experiment from our evaluation (Section~\ref{sec:evaluation}).

\section{Imagining identities with AC-GAN}
\label{sec:appendix_simplex}

\begin{figure}[H]
  \centering
  \includegraphics[width=0.96\linewidth]{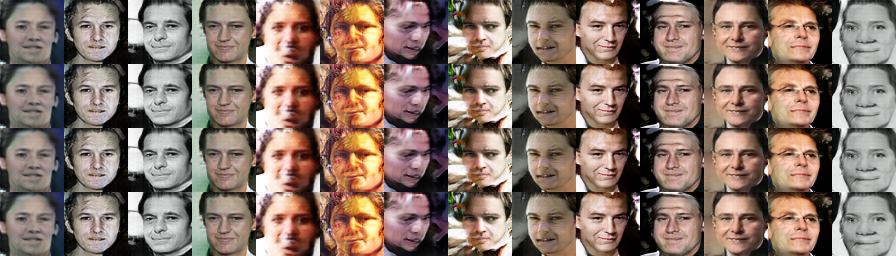}
  \caption{AC-DCGAN generation with random identity vectors that sum to one. Each row shares an identity vector and each column shares an observation vector.}
  \label{fig:acgan_simplex}
\end{figure}

\begin{figure}[H]
  \centering
  \includegraphics[width=0.96\linewidth]{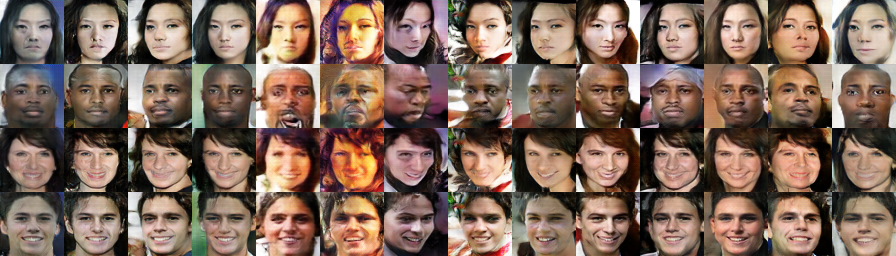}
  \caption{AC-DCGAN generation with one-hot identity vectors. Each row shares an identity vector and each column shares an observation vector.}
  \label{fig:acgan_reg}
\end{figure}

As stated in Section~\ref{sec:exp_deets}, 
AC-GANs~\citet{odena2016conditional} provide no obvious way to imagine new identities. 
For our evaluation (Section~\ref{sec:evaluation}), 
the AC-GAN generator receives identity input $\vect{z_I}~\in~[0, 1]^{10000}$: 
a one-hot over all identities. 
One possible approach to imagining \emph{new} identities would be to query a trained AC-GAN generator with a random vector $\vect{z_I}$ such that $\sum_{i=1}^{10000}~\vect{z_I[i]}~=~1$. 
We found that this strategy produced little identity variety (Figure~\ref{fig:acgan_simplex}) compared to the normal one-hot strategy (Figure~\ref{fig:acgan_reg}) and excluded it from our evaluation. 

\section{Architecture Descriptions}
\label{sec:appendix_arch}

We list here the full architectural details for our SD-DCGAN and SD-BEGAN models.
In these descriptions, $k$ is the number of images that the generator produces and discriminator observes per identity (usually $2$ for pairwise training), 
and $d_I$ is the number of dimensions in the latent space $\mathcal{Z}_I$ (identity). In our experiments, dimensionality of $\mathcal{Z}_O$ is always $100-d_I$. As a concrete example, the bottleneck layer of the SD-BEGAN discriminator autoencoder (``fc2'' in Table~\ref{tab:sdbegan_disc}) with $k=2,d_I=50$ has output dimensionality $150$.

We emphasize that generators are parameterized by $k$ in the tables only for clarity and symmetry with the discriminators. Implementations need not modify the generator; instead, $k$ can be collapsed into the batch size.

For the stacked-channels versions of these discriminators, we simply change the number of input image channels from $3$ to $3k$ and set $k=1$ wherever $k$ appears in the table.

\clearpage

\begin{table}[H]
\caption{Input abstraction for both SD-DCGAN and SD-BEGAN generators during training (where $\vect{z}_O$ is always different for every pair or set of $k$)}
\centering
\footnotesize
\begin{tabular}{ l | l | l | l}
Operation & Input Shape & Kernel Size & Output Shape\\
\hline
\input{archtables/input.tex}
\end{tabular}
\end{table}

\begin{table}[H]
\caption{SD-DCGAN generator architecture}
\centering
\footnotesize
\begin{tabular}{ l | l | l | l}
Operation & Input Shape & Kernel Size & Output Shape\\
\hline
\input{archtables/iddcgan_g.tex}
\end{tabular}
\end{table}

\begin{table}[H]
\caption{SD-DCGAN discriminator architecture}
\centering
\footnotesize
\begin{tabular}{ l | l | l | l}
Operation & Input Shape & Kernel Size & Output Shape\\
\hline
\input{archtables/iddcgan_d.tex}
\end{tabular}
\label{tab:dcgan_disc}
\end{table}

\begin{table}[H]
\caption{SD-BEGAN generator architecture}
\centering
\footnotesize
\begin{tabular}{ l | l | l | l}
Operation & Input Shape & Kernel Size & Output Shape\\
\hline
\input{archtables/idbegan_g.tex}
\end{tabular}
\label{tab:sdbegan_genr}
\end{table}

\begin{table}[H]
\caption{SD-BEGAN discriminator autoencoder architecture. The decoder portion is equivalent to, but \textbf{does not share weights with}, the SD-BEGAN generator architecture (Table~\ref{tab:sdbegan_genr}).}
\centering
\footnotesize
\begin{tabular}{ l | l | l | l}
Operation & Input Shape & Kernel Size & Output Shape\\
\hline
\input{archtables/idbegan_d.tex}
\end{tabular}
\label{tab:sdbegan_disc}
\end{table}

\section{Face Samples}
\label{sec:honest_samples}

We present samples from each model reported in Table \ref{facepairs} for qualitative comparison. In each matrix, $\vect{z}_I$ is the same across all images in a row and $\vect{z}_O$ is the same across all images in a column. We draw identity and observation vectors randomly for these samples.

\begin{figure}[H]
  \centering
  \includegraphics[width=0.96\linewidth]{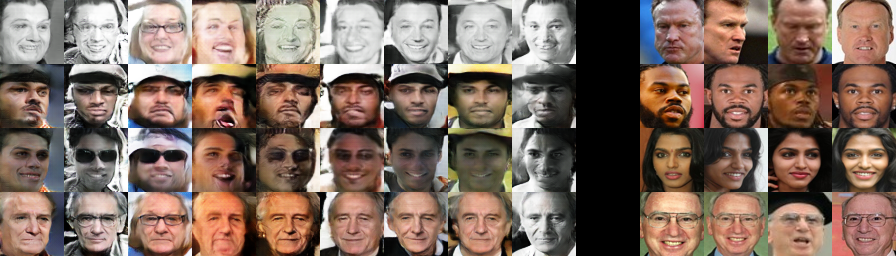}
  \caption{Generated samples from AC-DCGAN (four sample of real photos of ID on right)}
\end{figure}
\begin{figure}[H]
  \centering
  \includegraphics[width=0.96\linewidth]{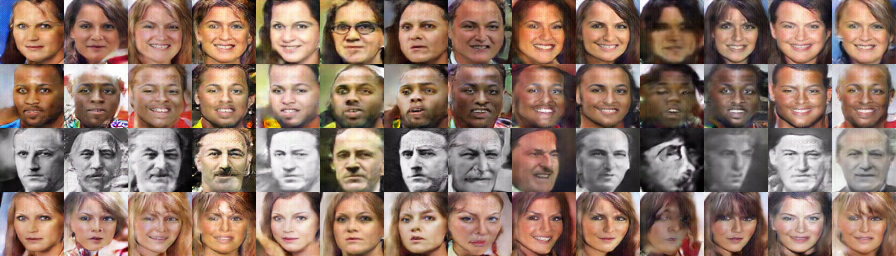}
  \caption{Generated samples from SD-DCGAN}
\end{figure}
\begin{figure}[H]
  \centering
  \includegraphics[width=0.96\linewidth]{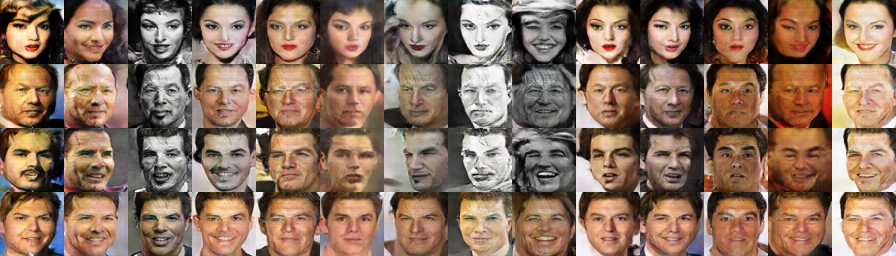}
  \caption{Generated samples from SD-DCGAN with stacked-channel discriminator}
\end{figure}
\begin{figure}[H]
  \centering
  \includegraphics[width=0.96\linewidth]{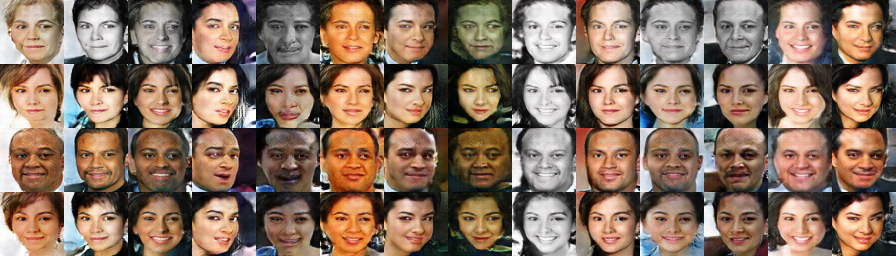}
  \caption{Generated samples from SD-DCGAN with $k=4$}
\end{figure}
\begin{figure}[H]
  \centering
  \includegraphics[width=0.96\linewidth]{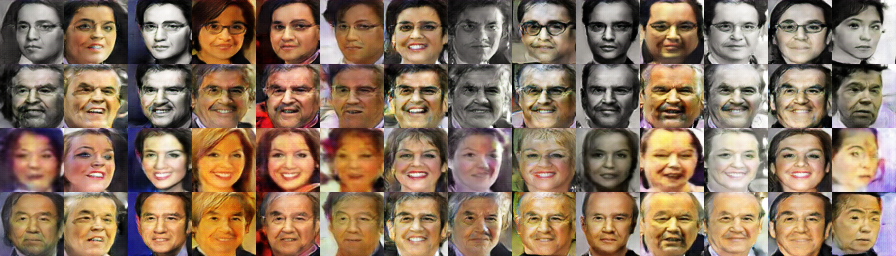}
  \caption{Generated samples from SD-DCGAN with $d_I=25$}
\end{figure}
\begin{figure}[H]
  \centering
  \includegraphics[width=0.96\linewidth]{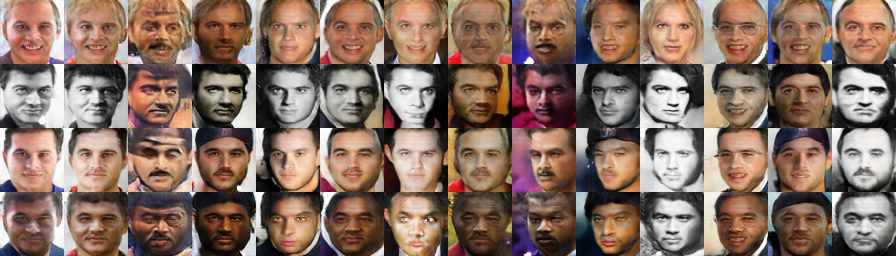}
  \caption{Generated samples from SD-DCGAN with $d_I=75$}
\end{figure}
\begin{figure}[H]
  \centering
  \includegraphics[width=0.96\linewidth]{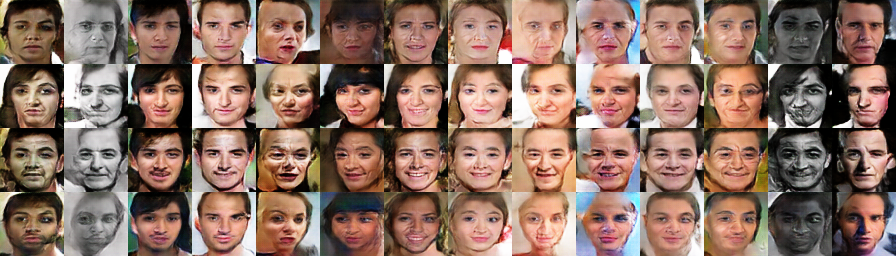}
  \caption{
  Generated samples from SD-DCGAN trained with the Wasserstein GAN loss~\citep{arjovsky2017wasserstein}.
  This model was optimized using RMS-prop~\citep{hinton2012neural} with $\alpha = 5\mathrm{e}{-5}$. 
  In our evaluation (Section~\ref{sec:evaluation}), 
  FaceNet had an AUC of $.770$ and an accuracy of $68.5$\% (at $\tau_v$) on data generated by this model. 
  We excluded it from Table~\ref{facepairs} for brevity.
  }
\end{figure}
\begin{figure}[H]
  \centering
  \includegraphics[width=0.96\linewidth]{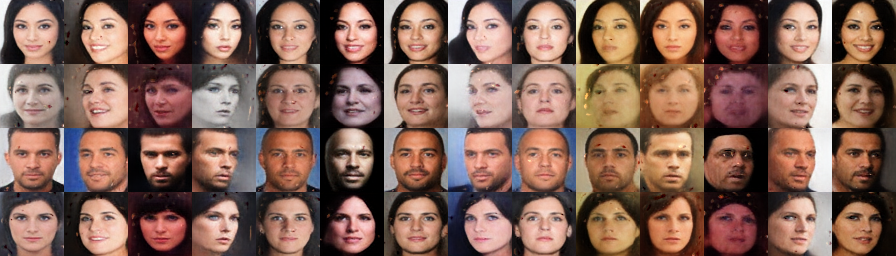}
  \caption{Generated samples from SD-BEGAN}
\end{figure}
\begin{figure}[H]
  \centering
  \includegraphics[width=0.96\linewidth]{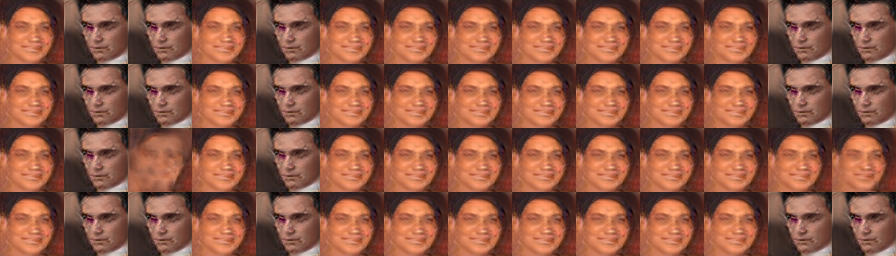}
  \caption{Generated samples from SD-BEGAN with $k=4$, demonstrating mode collapse}
\end{figure}

\section{Shoe Samples}
\label{sec:appendix_shoes}

We present samples from an SD-DCGAN and SD-BEGAN trained on our shoes dataset.

\begin{figure}[H]
  \centering
  \includegraphics[width=0.96\linewidth]{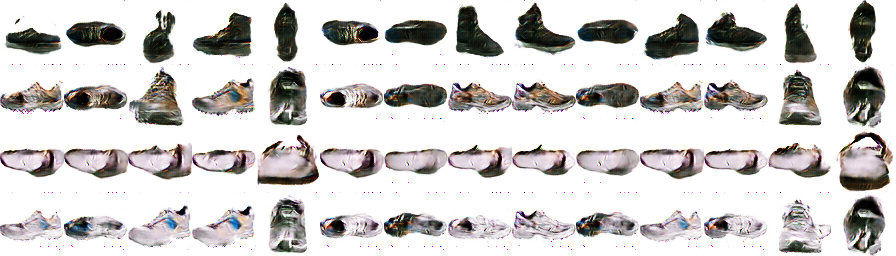}
  \caption{Generated samples from SD-DCGAN}
\end{figure}
\begin{figure}[H]
  \centering
  \includegraphics[width=0.96\linewidth]{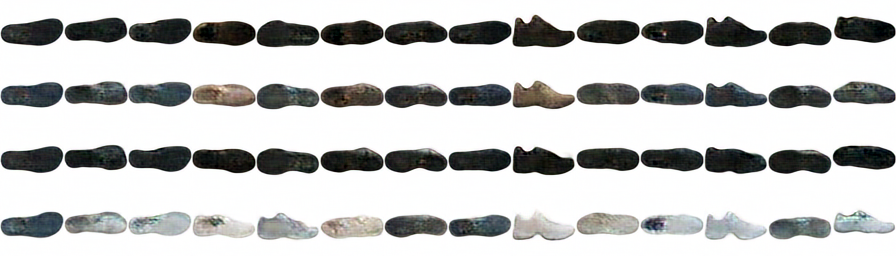}
  \caption{Generated samples from SD-BEGAN}
\end{figure}

%% file: archtables/input.tex
[$\boldsymbol{z}_i$; $\boldsymbol{z}_o$] & [($d_I$,);($k$,100-$d_I$)] &  & [($d_I$,);($k$,100-$d_I$)]\\
dup $\boldsymbol{z}_i$ & [($d_I$,);($k$,100-$d_I$)] &  & [($k$,$d_I$);($k$,100-$d_I$)]\\
concat & [($k$,$d_I$);($k$,100-$d_I$)] &  & ($k$,100)\\

%% file: archtables/iddcgan_g.tex
$\boldsymbol{z}$ & ($k$,100) &  & ($k$,100)\\
fc1 & ($k$,8192) & (100,8192) & ($k$,8192)\\
reshape & ($k$,8192) &  & ($k$,4,4,512)\\
bnorm & ($k$,4,4,512) &  & ($k$,4,4,512)\\
relu & ($k$,4,4,512) &  & ($k$,4,4,512)\\
upconv1 & ($k$,4,4,512) & (5,5,512,256) & ($k$,8,8,256)\\
bnorm & ($k$,8,8,256) &  & ($k$,8,8,256)\\
relu & ($k$,8,8,256) &  & ($k$,8,8,256)\\
upconv2 & ($k$,8,8,256) & (5,5,256,128) & ($k$,16,16,128)\\
bnorm & ($k$,16,16,128) &  & ($k$,16,16,128)\\
relu & ($k$,16,16,128) &  & ($k$,16,16,128)\\
upconv3 & ($k$,16,16,128) & (5,5,128,64) & ($k$,32,32,64)\\
bnorm & ($k$,32,32,64) &  & ($k$,32,32,64)\\
relu & ($k$,32,32,64) &  & ($k$,32,32,64)\\
upconv4 & ($k$,32,32,64) & (5,5,64,3) & ($k$,64,64,3)\\
tanh & ($k$,64,64,3) &  & ($k$,64,64,3)\\

%% file: archtables/iddcgan_d.tex
$\boldsymbol{x}$ or $G(\boldsymbol{z})$ & ($k$,64,64,3) &  & ($k$,64,64,3)\\
downconv1 & ($k$,64,64,3) & (5,5,3,64) & ($k$,32,32,64)\\
lrelu(a=0.2) & ($k$,32,32,64) &  & ($k$,32,32,64)\\
downconv2 & ($k$,32,32,64) & (5,5,64,128) & ($k$,16,16,128)\\
bnorm & ($k$,16,16,128) &  & ($k$,16,16,128)\\
lrelu(a=0.2) & ($k$,16,16,128) &  & ($k$,16,16,128)\\
downconv3 & ($k$,16,16,128) & (5,5,128,256) & ($k$,8,8,256)\\
bnorm & ($k$,8,8,256) &  & ($k$,8,8,256)\\
lrelu(a=0.2) & ($k$,8,8,256) &  & ($k$,8,8,256)\\
downconv4 & ($k$,8,8,256) & (5,5,256,512) & ($k$,4,4,512)\\
stackchannels & ($k$,4,4,512) &  & (4,4,512$k$)\\
downconv5 & (4,4,512$k$) & (3,3,512$k$,512) & (2,2,512)\\
flatten & (2,2,512) &  & (2048,)\\
fc1 & (2048,) & (2048,1) & (1,)\\
sigmoid & (1,) &  & (1,)\\

%% file: archtables/idbegan_g.tex
$\boldsymbol{z}$ & ($k$,100) &  & ($k$,100)\\
fc1 & ($k$,100,) & (100,8192) & ($k$,100,8192)\\
reshape & ($k$,100,8192) &  & ($k$,8,8,128)\\
conv2d & ($k$,8,8,128) & (3,3,128,128) & ($k$,8,8,128)\\
elu & ($k$,8,8,128) &  & ($k$,8,8,128)\\
conv2d & ($k$,8,8,128) & (3,3,128,128) & ($k$,8,8,128)\\
elu & ($k$,8,8,128) &  & ($k$,8,8,128)\\
upsample2 & ($k$,8,8,128) &  & ($k$,16,16,128)\\
conv2d & ($k$,16,16,128) & (3,3,128,128) & ($k$,16,16,128)\\
elu & ($k$,16,16,128) &  & ($k$,16,16,128)\\
conv2d & ($k$,16,16,128) & (3,3,128,128) & ($k$,16,16,128)\\
elu & ($k$,16,16,128) &  & ($k$,16,16,128)\\
upsample2 & ($k$,16,16,128) &  & ($k$,32,32,128)\\
conv2d & ($k$,32,32,128) & (3,3,128,128) & ($k$,32,32,128)\\
elu & ($k$,32,32,128) &  & ($k$,32,32,128)\\
conv2d & ($k$,32,32,128) & (3,3,128,128) & ($k$,32,32,128)\\
elu & ($k$,32,32,128) &  & ($k$,32,32,128)\\
upsample2 & ($k$,32,32,128) &  & ($k$,64,64,128)\\
conv2d & ($k$,64,64,128) & (3,3,128,128) & ($k$,64,64,128)\\
elu & ($k$,64,64,128) &  & ($k$,64,64,128)\\
conv2d & ($k$,64,64,128) & (3,3,128,128) & ($k$,64,64,128)\\
elu & ($k$,64,64,128) &  & ($k$,64,64,128)\\
conv2d & ($k$,64,64,128) & (3,3,128,3) & ($k$,64,64,3) \\

%% file: archtables/idbegan_d.tex
$\boldsymbol{x}$ or $G(\boldsymbol{z})$ & ($k$,64,64,3) &  & ($k$,64,64,3)\\
conv2d & ($k$,64,64,3) & (3,3,3,128) & ($k$,64,64,128)\\
elu & ($k$,64,64,128) &  & ($k$,64,64,128)\\
conv2d & ($k$,64,64,128) & (3,3,128,128) & ($k$,64,64,128)\\
elu & ($k$,64,64,128) &  & ($k$,64,64,128)\\
conv2d & ($k$,64,64,128) & (3,3,128,128) & ($k$,64,64,128)\\
elu & ($k$,64,64,128) &  & ($k$,64,64,128)\\
downconv2d & ($k$,64,64,128) & (3,3,128,256) & ($k$,32,32,256)\\
elu & ($k$,32,32,256) &  & ($k$,32,32,256)\\
conv2d & ($k$,32,32,256) & (3,3,256,256) & ($k$,32,32,256)\\
elu & ($k$,32,32,256) &  & ($k$,32,32,256)\\
conv2d & ($k$,32,32,256) & (3,3,256,256) & ($k$,32,32,256)\\
elu & ($k$,32,32,256) &  & ($k$,32,32,256)\\
downconv2d & ($k$,32,32,256) & (3,3,256,384) & ($k$,16,16,384)\\
elu & ($k$,16,16,384) &  & ($k$,16,16,384)\\
conv2d & ($k$,16,16,384) & (3,3,384,384) & ($k$,16,16,384)\\
elu & ($k$,16,16,384) &  & ($k$,16,16,384)\\
conv2d & ($k$,16,16,384) & (3,3,384,384) & ($k$,16,16,384)\\
elu & ($k$,16,16,384) &  & ($k$,16,16,384)\\
downconv2d & ($k$,16,16,384) & (3,3,384,512) & ($k$,8,8,512)\\
elu & ($k$,8,8,512) &  & ($k$,8,8,512)\\
conv2d & ($k$,8,8,512) & (3,3,512,512) & ($k$,8,8,512)\\
elu & ($k$,8,8,512) &  & ($k$,8,8,512)\\
conv2d & ($k$,8,8,512) & (3,3,512,512) & ($k$,8,8,512)\\
elu & ($k$,8,8,512) &  & ($k$,8,8,512)\\
flatten & ($k$,8,8,512) &  & ($k$,32768)\\
fc1 & ($k$,32768) & (32768,100) & ($k$,100)\\
concat & ($k$,100) &  & (100$k$,)\\
fc2 & (100$k$,) & (100$k$,$d_I$+(100-$d_I$)$k$,) & ($d_I$+(100-$d_I$)$k$,)\\
fc3 & ($d_I$+(100-$d_I$)$k$,) & ($d_I$+(100-$d_I$)$k$,100$k$,) & (100$k$,)\\
split & (100$k$,) &  & ($k$,100)\\
$G$ (Table~\ref{tab:sdbegan_genr}) & ($k$,100) &  & ($k$,64,64,3)\\